\documentclass{article}
\usepackage[utf8x]{inputenc}
\usepackage{spconf,amsmath,graphicx}

\graphicspath{{figures/}}

\usepackage{cite}
\usepackage{amsmath,amssymb,amsfonts}
\usepackage{algorithmic}
\usepackage{graphicx}
\usepackage{tikz}
\usepackage{hyperref}
\usepackage{lipsum}
\usepackage{caption}
\usepackage{siunitx}
\usepackage{commath}

\usepackage{subfig}

\usepackage{textcomp}
\usepackage{xcolor}
\def\BibTeX{{\rm B\kern-.05em{\sc i\kern-.025em b}\kern-.08em
    T\kern-.1667em\lower.7ex\hbox{E}\kern-.125emX}}

\begin{document}

\title{DFPN: Deformable Frame Prediction Network}

\name{M. Akın Yılmaz and A. Murat Tekalp 
\thanks{This work was supported by TUBITAK projects 217E033 and 120C156, and a grant from Turkish Is Bank to KUIS AILab. A. M. Tekalp also acknowledges support from Turkish Academy of Sciences (TUBA).}}

\address{Dept. of Electrical \& Electronics Eng., Koç University, 34450 İstanbul, Turkey \\
mustafaakinyilmaz@ku.edu.tr, mtekalp@ku.edu.tr }

\maketitle

\begin{abstract}
Learned frame prediction is a current problem of interest in computer vision and video processing/compression. Although several deep network architectures have been proposed for learned frame prediction, to the best of our knowledge, there is no work based on using deformable convolutions for frame prediction. To this effect, we propose a deformable frame prediction network (DFPN) for task-oriented implicit motion modeling and next frame prediction. Experimental results demonstrate that the proposed DFPN model achieves state of the art results in next frame prediction in sequences with global motion. Our models and results are available at \url{https://github.com/makinyilmaz/DFPN}.
\end{abstract}

\vspace{2pt}

\section{Introduction}
\label{intro}
Video frame prediction refers to predicting a future frame in a video given the current and past frames. It has been actively investigated in the fields of image/video processing and computer vision using deep learning. Although video frame prediction is formulated as a supervised learning problem, it can also be  considered as an unsupervised learning problem, since the ground-truth frames for frames to be predicted are already available in the video sequence itself. 

Main sources of motion in a video are camera motion and object motion, which are predictable to a certain extent by properly blending motion information from the past frames. However, predicting next frame in scenes having complex motion with both fast and slow moving objects in multiple directions can be difficult. Changes in object motion, camera pan angle changes, and sudden scene changes further complicate the problem.

Applications of frame prediction include autonomous driving systems, where it is vital to predict potential treats to take correct actions in time.
Another potential application of frame prediction is in video compression. 
Both classical or deep learning based video compression methods benefit from motion compensation using motion estimated between the current and reference frames~\cite{x264,x265,dvc,gop_compress}. However, estimated motion needs to be sent as side information increasing the bitrate.
If the future frame can be predicted from past frames fairly well, the prediction error can be coded at very low bitrates without a need for sending motion as side information. A video codec based on this concept was presented in~\cite{serkan}. 

\begin{figure}[t]
\centering
	\includegraphics[scale=0.30]{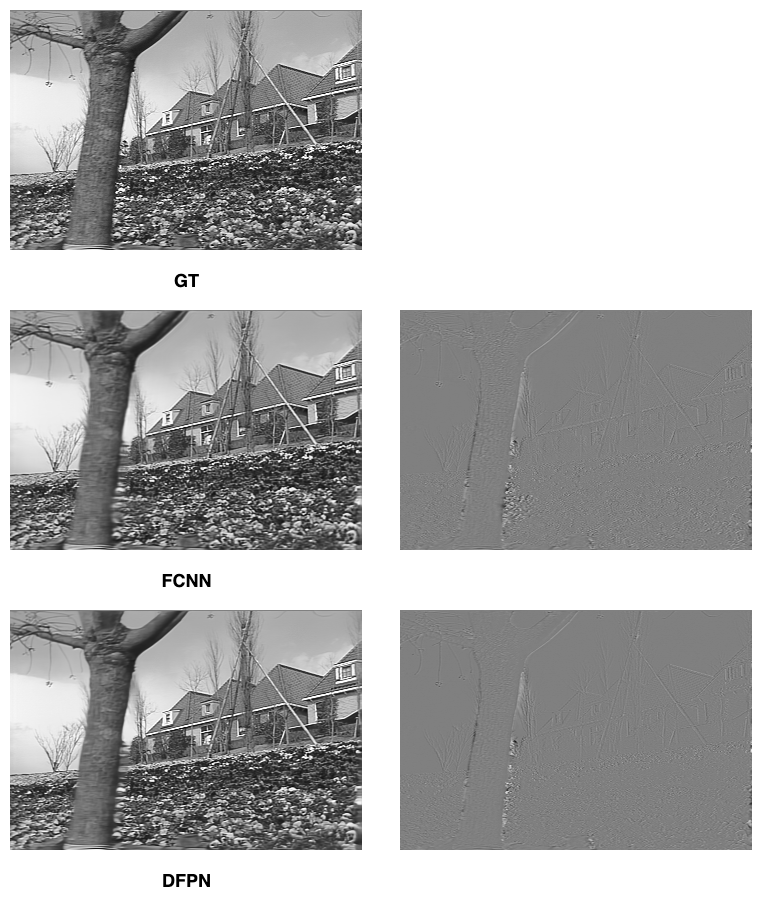} \vspace{-8pt} \\
\caption{Prediction results for a frame of \textit{Garden} sequence. PSNR - FCNN\cite{fcnn}: 25.75 dB, DFPN: 27.07 dB. Error images show DFPN gives better prediction near edges of the tree.} \vspace{-10pt}
\label{fig:residual}
\end{figure}

 In this paper, we propose a deformable frame prediction network (DFPN) for next frame prediction. Recent related works on frame prediction are reviewed in Section~\ref{related}. To the best of our knowledge, there is no work that uses deformable convolution for frame prediction. To this effect, we propose a deformable frame prediction network (DFPN) for task oriented implicit motion modeling. The proposed method is presented in Section~\ref{method}. In Section~\ref{experiments} experimental setup and results are explained. Finally, Section~\ref{conclusions} concludes the paper.


\section{Related work and Contributions}
\label{related}
\subsection{\textbf{Frame Prediction}} \vspace{-2pt}
Comprehensive reviews on video frame prediction methods can be found in \cite{review1, review2}. Methods covered in these reviews can be classified according to their network architecture, prediction methodology, and loss function used in training.

In terms of network architecture, most previous methods use some form of recurrent convolutional encoder-decoder architecture\cite{shi, fcnn}. There are also some methods that use 3D convolutions for handling temporal information\cite{crevnet}. 

Several works formulate the frame prediction problem as a synthesis problem directly in the pixel domain~\cite{shi, fcnn}, whereas others model similarity between successive frames by means of explicit transformations~\cite{fstn, sdcnet}.

Regarding optimization loss function, most works optimize $l_{1}$ or $l_{2}$ loss\cite{fcnn, sdcnet}. However, the resulting predicted frames may be blurry due to averaging effect. For more visually pleasing prediction results, some works combined perceptual or adversarial loss with the $l_{1}$ or $l_{2}$ loss\cite{sdcnet, villegas}.

\subsection{\textbf{Deformable Convolution}} \vspace{-2pt}
Deformable convolution was first proposed by~\cite{deform_cnn}. It is shown to be very effective in modeling geometric transformations between frames. 
It differs from the regular convolution by the 2D offsets that are added to the regular grid pixel sampling locations. 
The offsets enable deformable convolution to have larger and adaptive receptive field, which provides the ability to model complex and large motions. These offsets are learned during the training process and enable free form of deformations unlike optical flow based warping. 

Although it was originally proposed for object detection, recently it has also been used in the literature for frame alignment in video restoration and super resolution\cite{tdan, edvr}, where proper temporal alignment is crucial. When performing frame alignment using optical flow, some artifacts may be observed in the reconstructed frame due to errors in optical flow estimation and sub-pixel warping. Since deformable convolution performs one-stage alignment by implicit motion modelling in the feature space, it alleviates these adverse issues.
\vspace{-4pt}

\subsection{\textbf{Attention Mechanism}} \vspace{-2pt}
Attention mechanism allows the network to focus on the most informative parts of each frame for a specific task. Initial efforts to combine deep convolutional networks and attention mechanism focused on computer vision tasks such as image classification or object detection~\cite{ran,cbam}. In~\cite{cbam}, authors proposed convolutional block attention module (CBAM), which consists of separate channel and spatial attention blocks. They showed sequential stacking of these blocks gives better results compared to parallel stacking on object detection and image classification problems. CBAM was later applied to image denoising in~\cite{grdn}, which inspired us to use it in this work.
\vspace{-4pt}

\subsection{\textbf{Contributions}}
In our earlier work~\cite{fcnn}, we evaluated the performances of a recurrent convolutional-LSTM network and a fully convolutional network for pixel-domain frame prediction and found that the fully convolutional network provides better results. However, our convolutional network in~\cite{fcnn} used regular convolutional residual blocks. 

In this paper, we propose a deformable frame prediction network (DFPN) for task oriented implicit motion modeling and next frame prediction.
To the best of our knowledge, this is the first paper to propose a network using deformable convolutions for next frame prediction.
In addition to being a very lightweight model compared to~\cite{fcnn}, it is superior in terms of prediction performance.


\section{Proposed Method}

\label{method}
\begin{figure}[t]
\centering
	\includegraphics[scale=0.42]{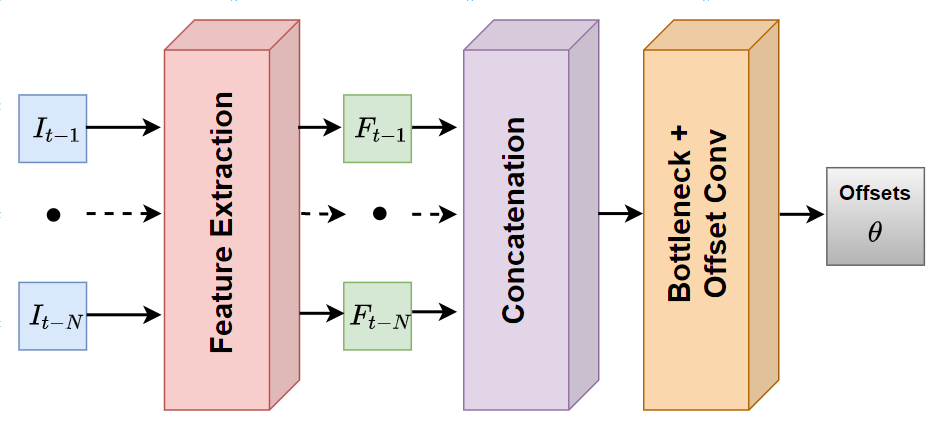} \vspace{-2pt} \\
	(a) \vspace{5pt}\\
	\includegraphics[scale=0.40]{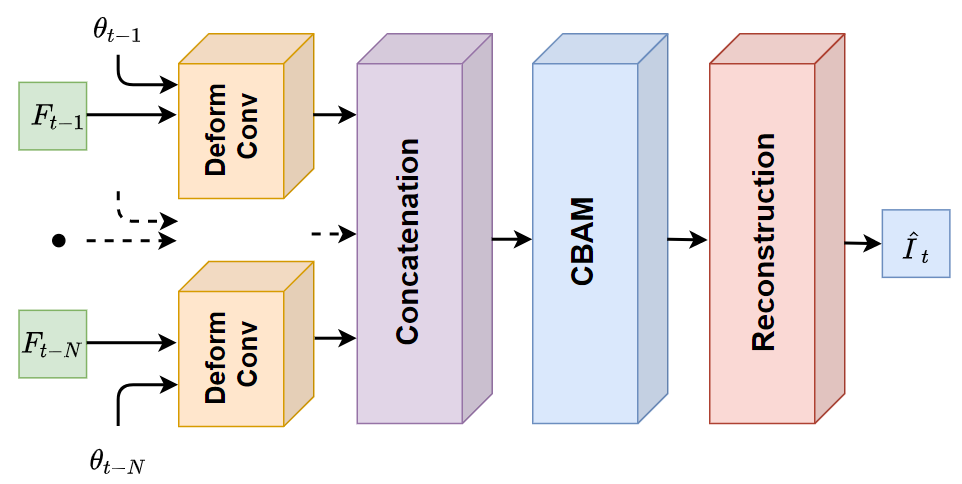} \vspace{-5pt} \\
	(b)
\caption{Our proposed DFPN framework: a) Feature extraction and offset prediction networks, (b) Predicted frame reconstruction network using deformable convolutions and offsets predicted in (a).}
\label{fig:our_model}
\end{figure}

Let $I_{t-N}, ..., I_{t-1} \in \mathbb{R^{H \times W \times C}}$ denote the last $N$ consecutive frames in a video sequence before the current frame $I_{t}$, which has the dimensions as previous frames. The~goal is to predict the current frame $I_{t}$ given previous frames $I_{t-N}, ..., I_{t-1}$. 

\subsection{\textbf{Overview of DFPN}}
The architecture of the proposed DFPN is depicted in Fig.~\ref{fig:our_model}. It consists of six sub-modules, which are detailed in the following. The feature extraction module extracts useful features of each input frame separately. Then, the bottleneck and offset convolution modules predict $N$ 2D offsets to perform $N$ deformable convolutions in the feature space. Note that, in the frame prediction problem, the current frame is not available as a reference frame for alignment. Therefore, we are not performing any alignment but we are predicting the offsets which implicitly predicts motion cues for predicting the next frame. After deformable convolutions, attention and reconstruction modules take deformed feature maps and fuse them to generate a predicted current frame. As the spatial dimensions remain unchanged during the forward pass, the model can process frames with arbitrary dimensions.

\subsection{\textbf{Feature Extraction}}
\begin{figure}[t]
\centering
	\includegraphics[scale=0.360]{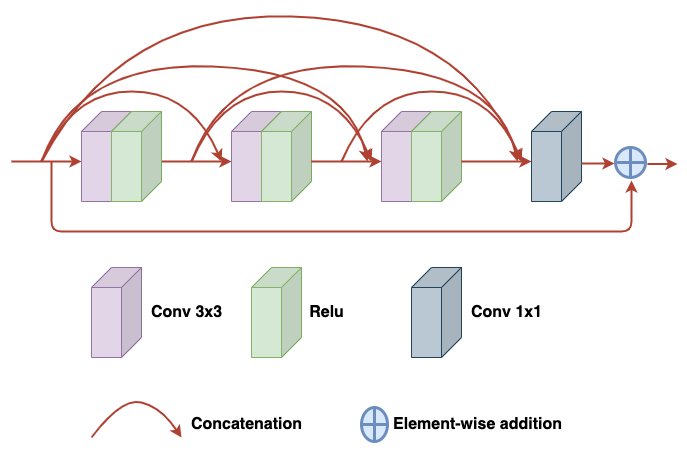} \vspace{-14pt} \\
\caption{Residual dense block architecture}
\label{fig:res_dense}
\end{figure}

Feature extraction is a shared module, which extracts features from $I_{t-N}, ..., I_{t-1}$ separately. It consists of one convolution layer, and 2 residual dense blocks~\cite{rdb} as shown in Figure~\ref{fig:res_dense} with 64 channels and ReLU activation. The extracted features $F_{t-N}, ..., F_{t-1}$ are passed to the bottleneck module.

\subsection{\textbf{Bottleneck and Offset Prediction}}
Bottleneck module takes channel-wise concatenated features and fuse them using 12 residual dense blocks. At the beginning of the module, one convolution operation is applied to reduce channel dimension by a factor of $N$ to have the same dimension as each extracted feature. 

At the end of bottleneck module, a convolution named offset convolution exists. It takes fused feature map with 64 channels, and predicts $N$ 2D offsets each has shape $H \times W \times 18 (2 \times 3 \times 3)$. The last dimension is 18 since deformable convolution kernel is set to $3 \times 3$. For stable training, offset convolution weights and biases are initialized to zeros, that is, no deformation is assumed at the initial state. 

\subsection{\textbf{Deformable Convolution}}
The deformable convolution module consists of $N$ separate layers. Each takes one of the individual features $F_{t-N}, ..., F_{t-1}$ and predicted offsets $\theta_{t-N}, ..., \theta_{t-1}$ and performs deformable convolution on feature level unlike optical flow based methods. More specifically, let $R = \{ (-1,-1),(-1,0),...,(1,1)\}$ denotes $3 \times 3$ kernel used to sample a region of the input feature map.  Then, the deformed feature map at each position $p_{0}$ can be computed as:
\begin{equation}
{y(p_{0})} = \sum_{p_{n}\in{R}} w(p_{n}) \cdot x(p_{0}+p_{n}+\Delta p_{n})
\label{deformable_conv}
\end{equation}
In the normal convolution operation $\Delta p_{n}$ is zero and the sampling grid is regular. Deformable convolution operation works on irregular sampling locations specified by offsets $\Delta p_{n}$. These sampling locations are generally subpixel and bilinear interpolation is performed to compute sub-pixel intensities~\cite{deform_cnn}.

\subsection{\textbf{Attention}}
After deformable convolution operations, convolutional block attention module~\cite{cbam} takes concatenated feature maps, and returns refined version of features. This process covers separate channel and/or spatial attentions applied in sequential order. Channel attention is designed to focus on the most informative parts of channel dimension using global average and max pooling operations applied on spatial dimensions. For spatial attention, pooling operations are applied on the channel dimension to get the spatial attention weights. 

\subsection{\textbf{Reconstruction}}
Reconstruction module consists of 8 residual dense blocks with 64 input channels and one convolution layer at the end to reduce channel size to the number of image channels (one for gray-scale prediction). It takes concatenated deformed features of $N$ previous frames with or without attention mechanism. The module directly estimates pixels of the next frame in a video sequence.


\section{Experiments}
\label{experiments}

\begin{figure*}[t]
\centering
\subfloat      {
\includegraphics[width=.246\linewidth]{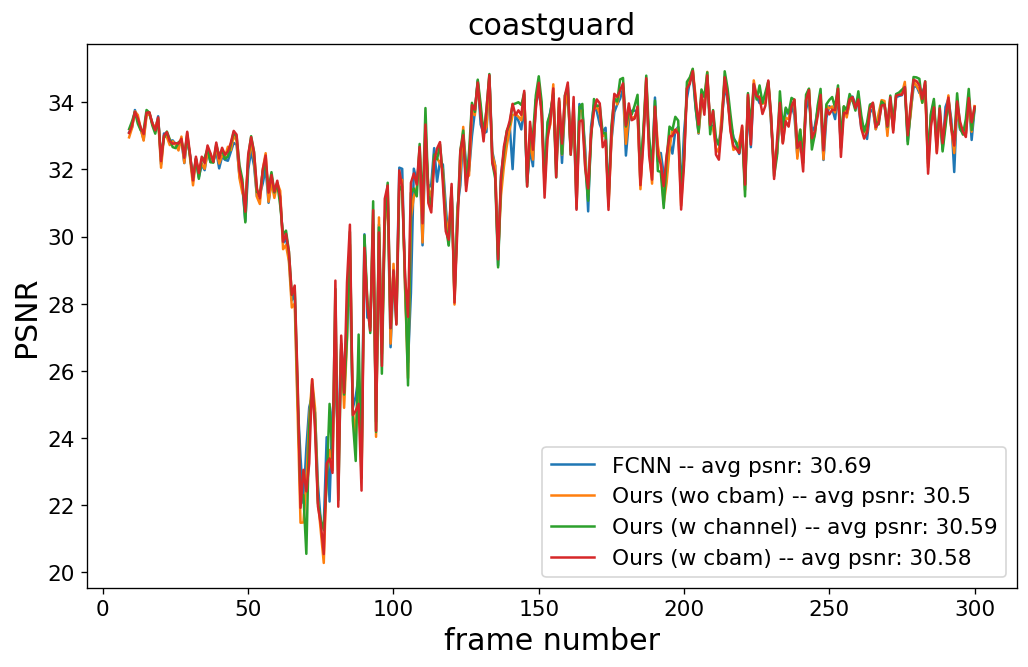}
} \hspace{-8pt}
\subfloat     {
\includegraphics[width=.246\linewidth]{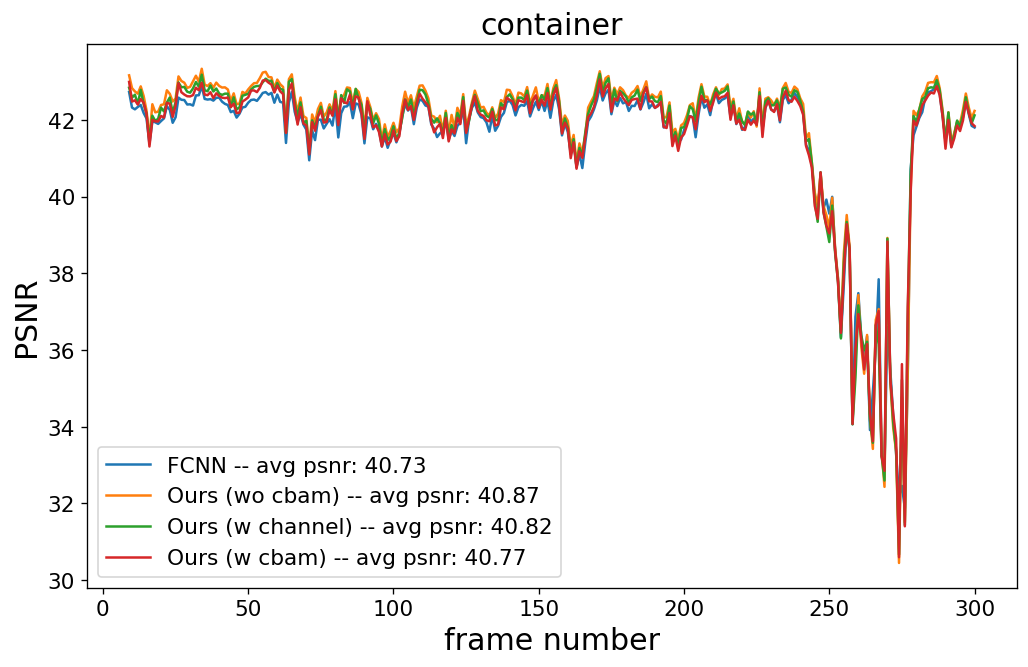}
} \hspace{-8pt}
\subfloat     {
\includegraphics[width=.246\linewidth]{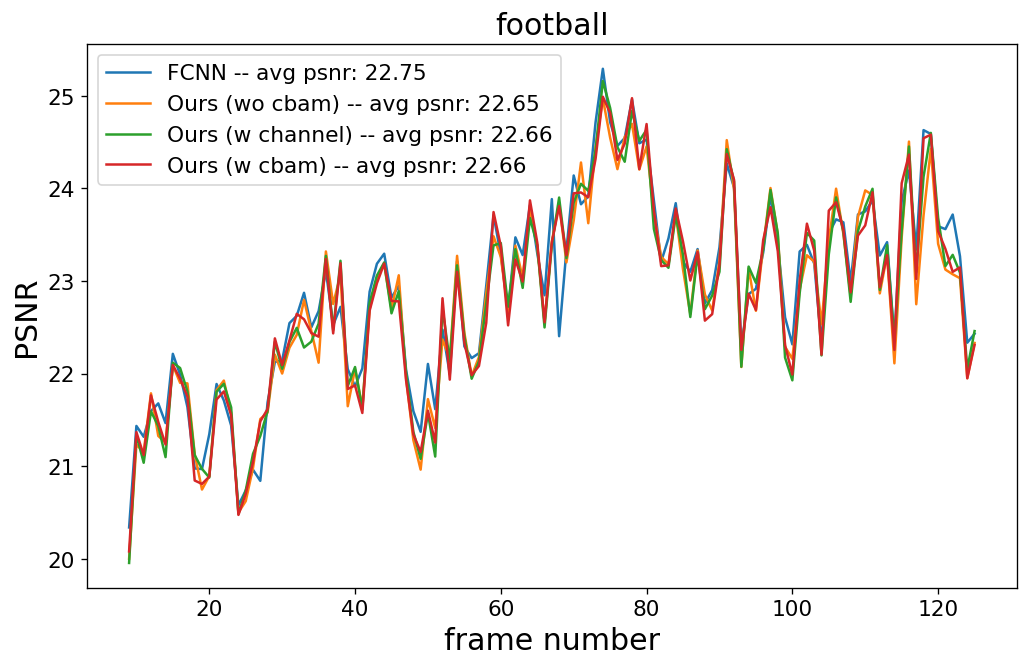}
} \hspace{-8pt} 
\subfloat     {
\includegraphics[width=.246\linewidth]{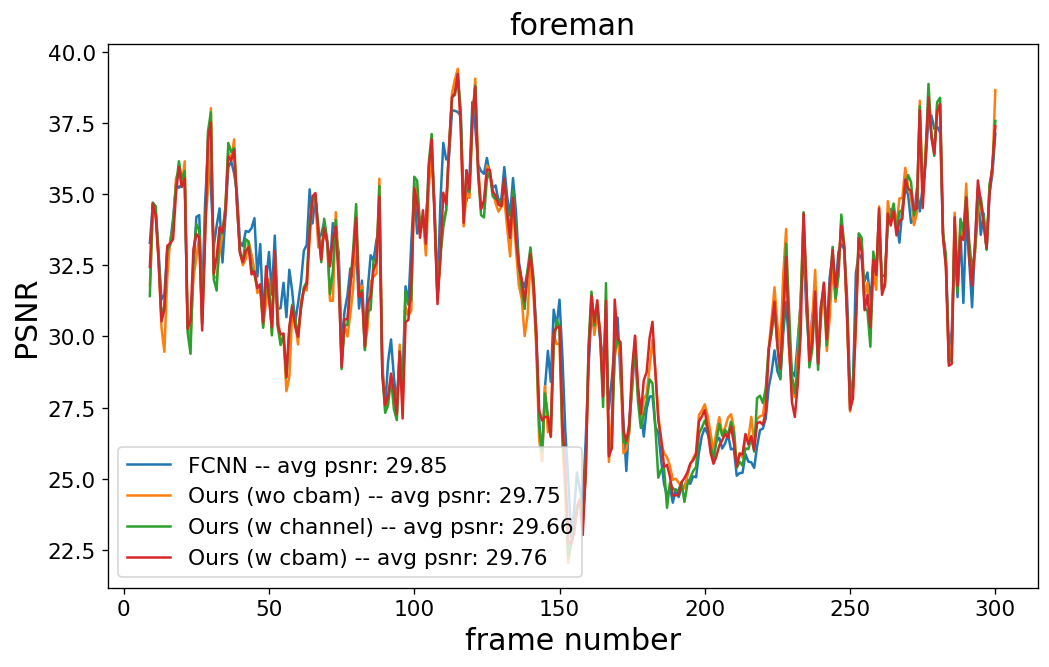}
} \vspace{-5pt} \\
\subfloat     {
\includegraphics[width=.246\linewidth]{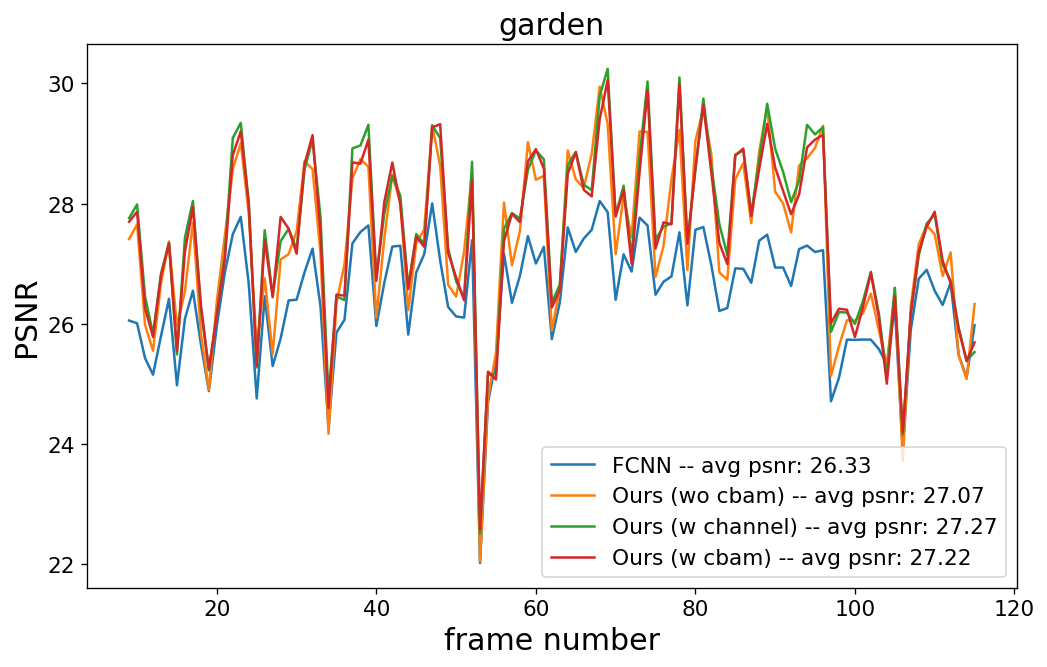}
} \hspace{-8pt}
\subfloat     {
\includegraphics[width=.246\linewidth]{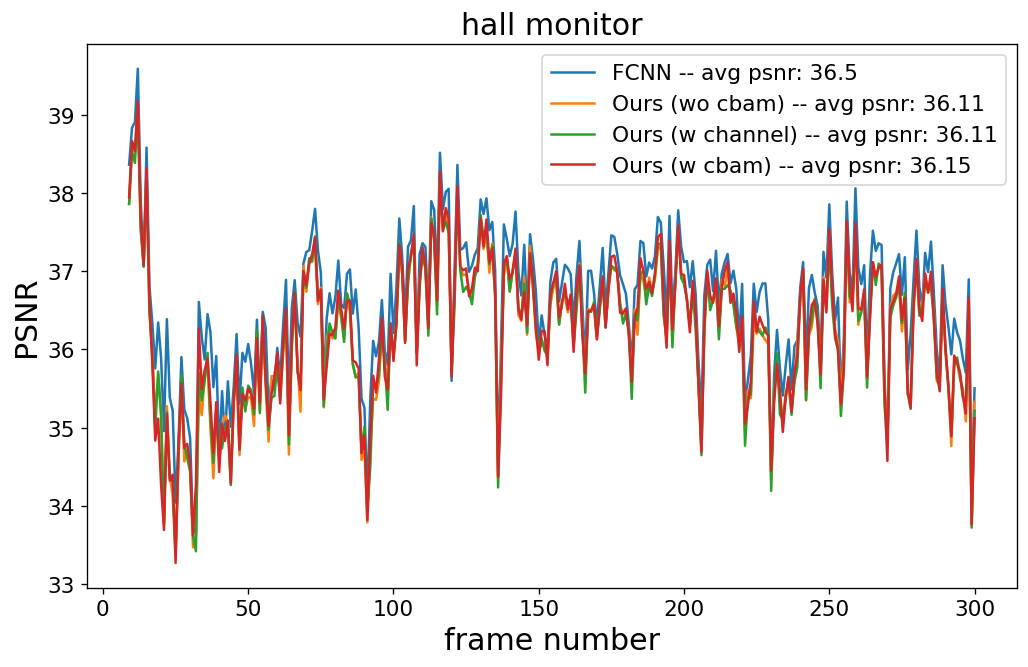}
}\hspace{-8pt}
\subfloat     {
\includegraphics[width=.246\linewidth]{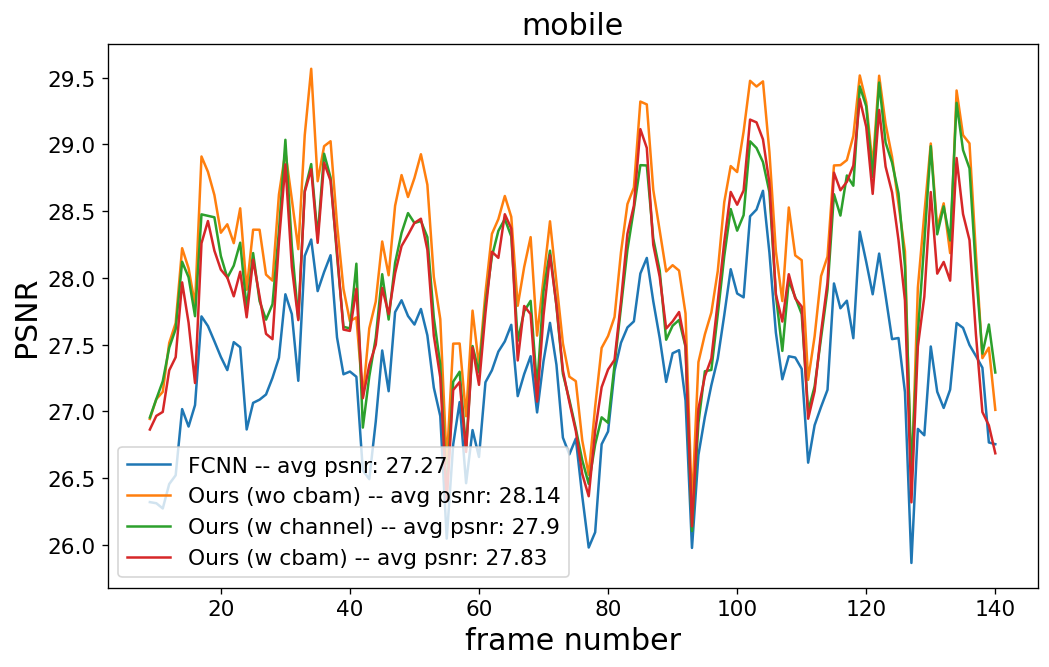}
} \hspace{-8pt}
\subfloat     {
\includegraphics[width=.246\linewidth]{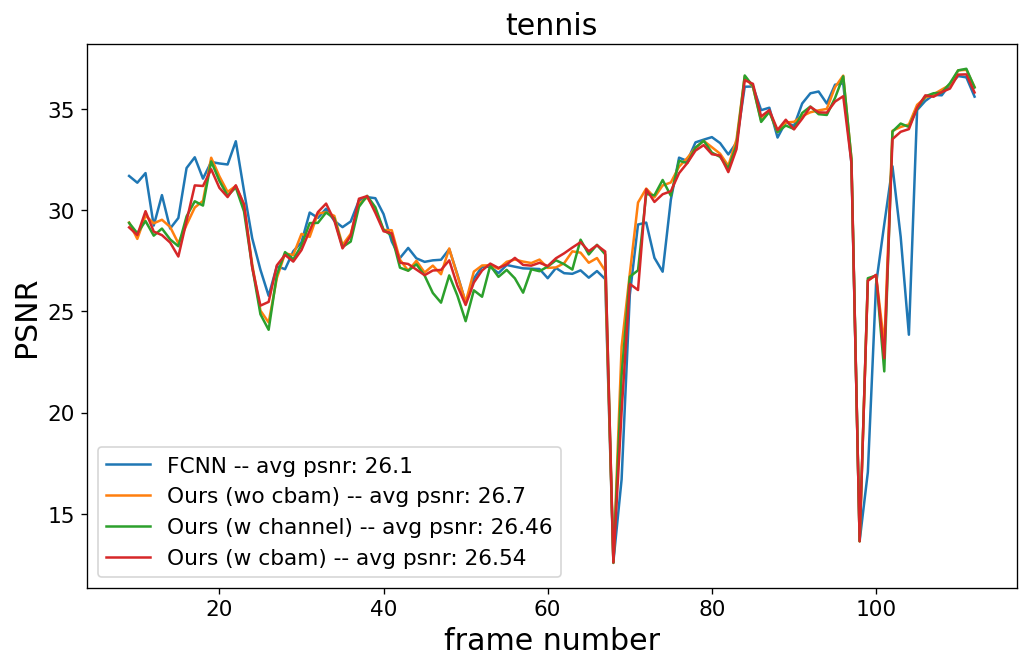}
} \vspace{2pt}\\
\caption{Comparison of prediction PSNR between DFPN w/wo attention and FCNN\cite{fcnn} vs. frame number on MPEG test videos.}
\label{fig:prediction}
\end{figure*}

\subsection{Experimental Setup}
\subsubsection{\textbf{Training Settings}}
We train our deformable frame prediction network on gray-scale frames extracted from UCF101 dataset\cite{ucf101}. We randomly select 5 consecutive frames and apply $96 \times 96$ random cropping to the same positions in order to create an augmented batch of consecutive patches. The first 4 frames of a batch are used as input to our model and the 5th frame is taken as the ground-truth frame. We optimize $L_{1}$ loss, given by
\begin{equation}
L_{1} = \norm{I_{gt}-I_{pred}}_1
\label{loss_equation}
\end{equation}
during training since we observed that it produces less blurry images compared to the $L_{2}$ loss. We set the batch size as 8 and used Adam\cite{adam} optimizer for 500K iterations. Initial learning rate is set to 0.0001 and halved in every 100K iterations.

\subsubsection{\textbf{Test Sequences and Evaluation}}
We test the performance of our model on 8 MPEG sequences: Coastguard, Container, Football, Foreman, Garden, Hall Monitor, Mobile, and Tennis and compare the results with those in \cite{fcnn}. PSNR is used for our evaluation metric.

\subsection{Experimental Results}

In this section, we present quantitative evaluation of prediction performances. To this effect, graphs of PSNR vs. frame number are plotted for 8 MPEG sequences in Figure~\ref{fig:prediction}.

Inspection of Figure~\ref{fig:prediction} reveals that the proposed DFPN with or without attention gives superior or very close results to those in ~\cite{fcnn} for all sequences. This indicates that it is more effective to learn motion cues using deformable convolutions in the feature space compared to using regular convolutions on channel-wise stacked input image frames as in~\cite{fcnn}. 

If a more detailed video-based review is to be made, for videos containing global camera motion like \textit{garden} and \textit{mobile}, a huge performance gain can be observed in the favor of DFPN. An example frame from \textit{garden} sequence is shown in Figure~\ref{fig:residual} along with prediction results of DFPN and FCNN~\cite{fcnn}. DFPN method gives around 1.3 dB improvement in terms of PSNR compared to former method. 

In terms of model complexity and runtime, DFPN is a more lightweight network which has near 6 million parameters compared to \cite{fcnn} with over 38 million parameters. It was reported that FCNN can generate 1 frame per second on  a  single  NVIDIA  GeForce  GTX  1080Ti  GPU. However, DFPN can generate 4 frames per second on the same hardware. Although DFPN cannot make real-time estimation, its 4 times faster inference besides its prediction power, provides superiority to DFPN in terms of both performance and simplicity.

Although DFPN without any channel or spatial attention blocks already provides very good results, three experiments were conducted to investigate the effect of including a convolutional block attention module. Experiments indicate that models with channel and/or spatial attention blocks provides better results for some videos in terms of PSNR performance, while for some other videos no clear advantage of using channel attention blocks has been observed. 

\vspace{10pt}


\section{Conclusion}
\label{conclusions}
We propose a promising deformable frame prediction network (DFPN) that makes use of deformable convolutions for fusion of features from previous frames. Compared to the standard convolutional network approach FCNN~\cite{fcnn}, our method yields far better results in sequences with global motion and very close results in sequences with slow motion in terms of the average PSNR. Furthermore, the number of learned parameters in DFPN is one sixth of the number of parameters in FCNN. Since DFPN is a lighter model, it allows faster frame prediction inference.

To the best of our knowledge, this paper is the first to propose deformable convolutions for video frame prediction. As for future work, a multi-scale, coarse-to-fine pyramid structure, similar to the one used in EDVR~\cite{edvr}, can be incorporated into DFPN that should positively affect prediction performance. In addition, investigation of more sophisticated spatio-temporal attention modules can further improve the performance. 


\clearpage
\bibliography{references}
\bibliographystyle{IEEEtran}
\end{document}